\documentclass[runningheads]{llncs}

\usepackage{eccv}

\usepackage{eccvabbrv}

\usepackage{graphicx}
\usepackage{booktabs}

\usepackage[accsupp]{axessibility}  % Improves PDF readability for those with disabilities.

\usepackage{hyperref}

\usepackage{orcidlink}

\usepackage[linesnumbered,ruled,vlined]{algorithm2e}

\begin{document}

\title{AnchoredDream: Zero-Shot 360° \\Indoor Scene Generation from a Single View \\via Geometric Grounding} 

\titlerunning{AnchoredDream: Single View 360° Scene Generation}

\author{
Runmao Yao\inst{1}$^{*}$, 
Junsheng Zhou\inst{1}$^{*}$$^{\dagger}$,
Zhen Dong\inst{2},
Yu-Shen Liu\inst{1}$^{\dagger}$
}

\newcommand{\customfootnote}[2]{%
  \begingroup
  \renewcommand{\thefootnote}{#1}%
  \footnotetext{#2}%
  \endgroup
}

\customfootnote{$*$}{Equal contribution.}
\customfootnote{$\dagger$}{Corresponding author.}

\authorrunning{R. Yao, J. Zhou, Z. Dong, Y.-S. Liu}

\institute{School of Software, Tsinghua University \and Wuhan University \\
\email{\{yrm21, zhou-js24\}@mails.tsinghua.edu.cn} \\ 
\email{dongzhenwhu@whu.edu.cn}, \email{liuyushen@tsinghua.edu.cn}}

\maketitle

\begin{figure}[ht]
  \centering
  \includegraphics[width=\linewidth]{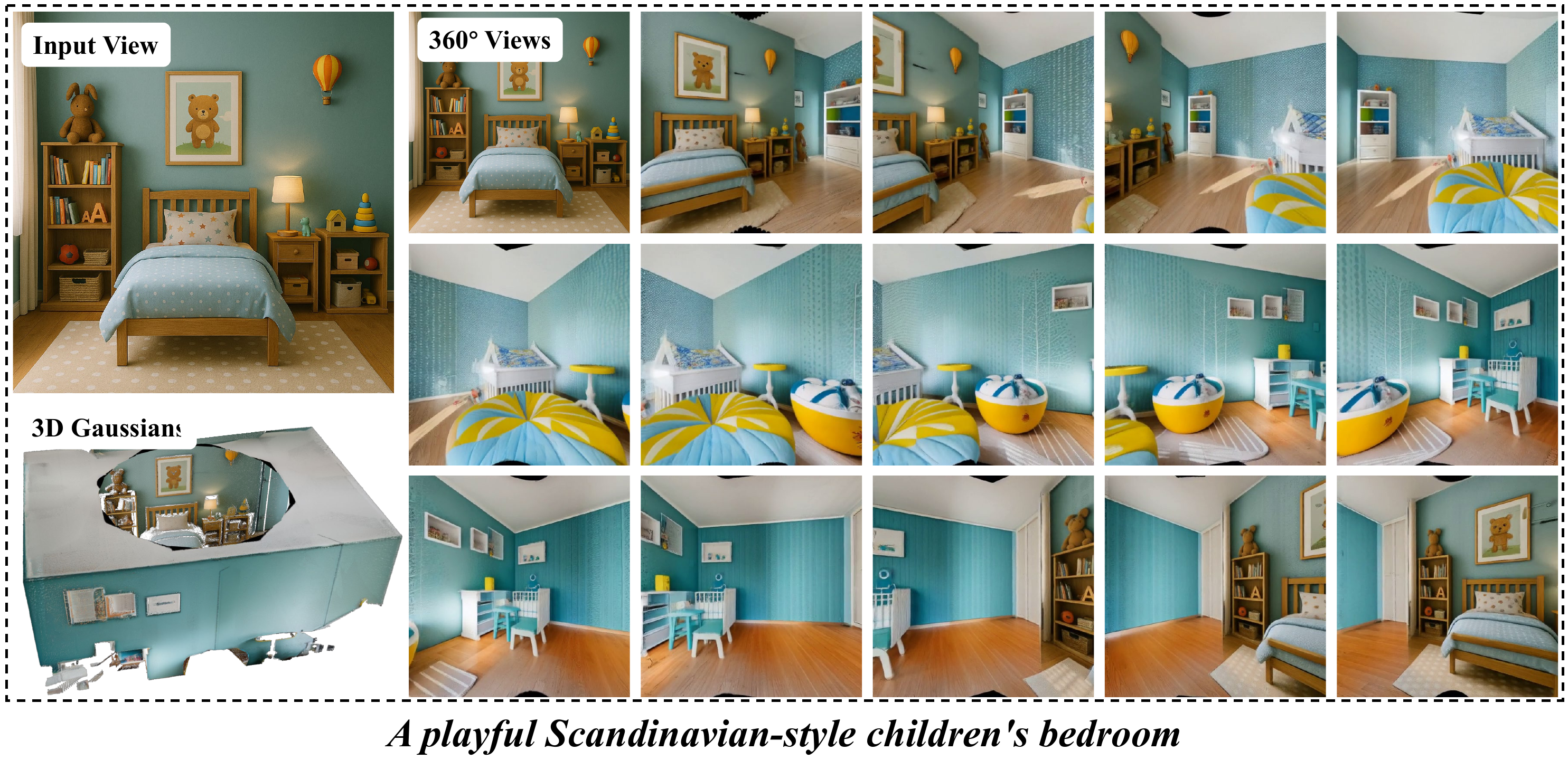}
  \caption{\textbf{Single-view to 360° scene generation.} Given a single-view indoor image, AnchoredDream generates a complete 360° scene represented with 3DGS, while maintaining both appearance consistency and geometric plausibility throughout.}
  \label{fig:demo}
  \vspace{-15pt}
\end{figure}

\begin{abstract}
  Single-view indoor scene generation plays a crucial role in a range of real-world applications. However, generating a complete 360° scene from a single image remains a highly ill-posed and challenging problem. Recent approaches have made progress by leveraging diffusion models and depth estimation networks, yet they still struggle to maintain appearance consistency and geometric plausibility under large viewpoint changes, limiting their effectiveness in full-scene generation. To address this, we propose \textbf{AnchoredDream}, a novel zero-shot pipeline that anchors 360° scene generation on high-fidelity geometry via an appearance-geometry mutual boosting mechanism. Given a single-view image, our method first performs appearance-guided geometry generation to construct a reliable 3D scene layout. Then, we progressively generate the complete scene through a series of modules: warp-and-inpaint, warp-and-refine, post-optimization, and a novel \textbf{Grouting Block}, which ensures seamless transitions between the input view and generated regions. Extensive experiments demonstrate that AnchoredDream outperforms existing methods by a large margin in both appearance consistency and geometric plausibility—all in a zero-shot manner. Our results highlight the potential of geometric grounding for high-quality, zero-shot single-view scene generation.
  \keywords{Single-View 3D Scene Generation \and Novel View Synthesis \and Gaussian Splatting}
\end{abstract}

\section{Introduction}

Generating indoor scenes from a single-view image has broad practical applications. For instance, online home furnishing platforms such as IKEA and Wayfair can automatically generate room layouts from a single user-provided image, enabling more accurate and personalized furniture recommendations. In indoor robot navigation, homecare robots can infer the full room structure from a forward-facing view, which allows them to gain a more comprehensive understanding of the environment. As a result, they can plan paths and execute tasks more effectively~\cite{Yao_2025_CVPR, zou2018layoutnet}.

However, generating a complete and coherent 360° scene from a single image remains a fundamentally challenging and ill-posed problem, as it requires strict consistency in both appearance and geometry. To address this challenge, recent methods~\cite{shriram2025realmdreamertextdriven3dscene, chung2023luciddreamerdomainfreegeneration3d, höllein2023text2roomextractingtextured3d, zhang2024text2nerftextdriven3dscene} aim to distill 2D priors, building upon the success of early works~\cite{gu2023nerfdiffsingleimageviewsynthesis, poole2022dreamfusiontextto3dusing2d, wang2023prolificdreamerhighfidelitydiversetextto3d, yi2023gaussiandreamer} in object-level novel view synthesis. Inevitably, these methods inherit a key limitation of 2D diffusion models: a lack of 3D perception~\cite{yoo2023dreamsparseescapingplatoscave}, which makes it difficult to maintain spatial consistency across views, especially under large viewpoint changes. Several approaches have attempted to address this challenge. For instance, CAT3D~\cite{gao2024cat3dcreate3dmultiview} improves multi-view consistency by fine-tuning diffusion models on multi-view datasets. FlexWorld~\cite{chen2025flexworldprogressivelyexpanding3d} further adopts video-to-video diffusion models trained on carefully designed data to enhance spatial coherence. Despite these efforts, large-scale high-quality 3D scene data is difficult to collect~\cite{hou2021exploringdataefficient3dscene}, which remains a major bottleneck that hinders further progress. To overcome this limitation, recent works explore warp-and-inpaint approaches. These methods~\cite{lei2023rgbd2generativescenesynthesis, wang2024vistadreamsamplingmultiviewconsistent} gradually synthesize novel views using monocular depth estimation (MDE)~\cite{bochkovskii2025depthprosharpmonocular, depth_anything_v1, depth_anything_v2} and image inpainting~\cite{lllyasviel2023fooocus}. However, they often suffer from noisy depth predictions and accumulated reprojection errors, resulting in unreliable warped images. As a result, the synthesized views are typically constrained to viewpoints near the input image. Even when these methods attempt to warp aggressively to cover a full 360° scene, the results often suffer from low appearance quality and inconsistent geometry.

After carefully analyzing previous approaches, we identify a key missing element: \textbf{reliable geometric grounding}. High-fidelity geometry inherently provides strong 3D perception, enhancing the reliability and consistency of 2D priors during scene generation. Additionally, precise geometry avoids error accumulation across viewpoint changes and remains robust under large parallax, offering a stable foundation for warping and subsequent inpainting. Moreover, geometry effectively guides the direction of scene generation. Therefore, grounding scene generation in accurate geometry is a highly promising solution.

Given this insight, we propose \textbf{AnchoredDream}, a novel pipeline that anchors 360° indoor scene generation on high-fidelity geometry through an appearance-geometry mutual boosting mechanism. Specifically, our pipeline involves the following key steps: first, given a single-view input image capturing partial scene appearance, we employ a vision-language model (VLM) to extract both extrinsic visual characteristics and intrinsic semantic context, generating an informative scene description. Next, guided by this description, we generate a 3D indoor scene layout, establishing a geometric scaffold that provides reliable spatial guidance for subsequent stages. Afterwards, conditioned on this scaffold, we progressively generate 360° appearance through warp-and-inpaint operations. Concurrently, we represent the generated scene as a set of 3D Gaussian primitives, enabling rendering at arbitrary novel views via Gaussian splatting~\cite{kerbl20233dgaussiansplattingrealtime}. To further enhance the appearance consistency across the generated scene, we employ warp-and-refine techniques, ensuring global stylistic coherence. Moreover, we introduce a novel \textbf{Grouting Block}, inspired by tile grouting in interior design, which enables mutual refinement between appearance and geometry, allowing the input view and the generated regions to blend seamlessly. Finally, we perform post-optimization to refine the 3D Gaussian representation, enhancing multi-view consistency and overall geometric plausibility. We summarize our main contributions as follows:

\begin{itemize}
    \item We propose AnchoredDream, a novel appearance-geometry mutual boosting pipeline that grounds scene appearance generation on high-fidelity geometry for 360° indoor scene generation from a single view.
    \item We design a novel Grouting Block to seamlessly blend the boundary between the input view and the generated regions, enhancing the consistency between scene appearances and geometry.
    \item Extensive experiments show that AnchoredDream outperforms existing methods in both appearance consistency and geometric plausibility in a zero-shot setting.
\end{itemize}

\section{Related Work}

\subsection{Layout Generation}

Layout generation originally began with 2D image layout generation~\cite{gupta2021layouttransformerlayoutgenerationcompletion, jyothi2021layoutvaestochasticscenelayout, kong2022bltbidirectionallayouttransformer, li2019layoutgangeneratinggraphiclayouts}. With the emergence of large language models (LLMs), LayoutGPT~\cite{feng2023layoutgptcompositionalvisualplanning} observed that image and indoor scene layouts share structural similarities and can be expressed in CSS-like formats, which are part of the LLM training corpus. This observation enabled them to bridge the domain gap with LLMs. However, directly predicting numerical values using LLMs often leads to layouts that violate physical plausibility. To address this issue, several approaches introduce intermediate representations to guide 3D layout generation~\cite{lin2024instructsceneinstructiondriven3dindoor, rahamim2024layascenepersonalized3dobject}. More recent methods~\cite{fu2024anyhomeopenvocabularygenerationstructured, littlefair2025flairgptrepurposingllmsinterior, pun2025hsmhierarchicalscenemotifs, wang2024architectgeneratingvividinteractive} focus on the integration of small objects into the scene to increase realism and functional diversity. In addition, other work~\cite{sun2025layoutvlmdifferentiableoptimization3d, yang2024holodecklanguageguidedgeneration} adopts an initialization-and-optimization paradigm, where object pose parameters are first initialized and then refined using predefined differentiable spatial constraints. Building on this, our method first generates indoor scene layouts and uses them as high-fidelity geometry to support downstream generation.

\subsection{Diffusion Models}

Diffusion models have emerged as a powerful class of generative models, initially introduced in the context of score-based generative modeling~\cite{sohldickstein2015deepunsupervisedlearningusing, song2021scorebasedgenerativemodelingstochastic}, and later popularized by the denoising diffusion probabilistic model framework~\cite{ho2020denoisingdiffusionprobabilisticmodels}. Their strong generation quality and stable training dynamics have led to widespread adoption across various tasks~\cite{dhariwal2021diffusionmodelsbeatgans, Du_2025_CVPR, saharia2022photorealistictexttoimagediffusionmodels}. To improve efficiency, latent diffusion models (LDMs)~\cite{rombach2022highresolutionimagesynthesislatent} were proposed, which operate in the latent space of pretrained autoencoders. This formulation significantly reduces computational requirements while retaining high image quality and flexibility, and was implemented in the widely-used Stable Diffusion~\cite{rombach2022highresolutionimagesynthesislatent} framework. In the context of multi-modal and conditional generation, diffusion models have been widely applied to tasks such as image inpainting~\cite{lugmayr2022repaintinpaintingusingdenoising} and conditional image synthesis~\cite{zhang2023addingconditionalcontroltexttoimage}. In our work, we leverage the depth-conditioned generation capabilities of diffusion models to produce a consistent scene appearance.

\subsection{Single-View 3D Scene Generation}

Single-view 3D scene generation methods often follow a common paradigm: distilling 2D priors into 3D scene generation, with diffusion models being the most frequently chosen prior. These methods can be broadly categorized into two classes: feed-forward and iterative optimization approaches. Feed-forward methods~\cite{chen2025flexworldprogressivelyexpanding3d, Chen_2023, shriram2025realmdreamertextdriven3dscene, tang2023mvdiffusionenablingholisticmultiview} typically adopt an end-to-end pipeline that generates the scene through a small number of forward passes. However, such approaches often require fine-tuning or large-scale training on 3D asset repositories~\cite{deitke2022objaverseuniverseannotated3d} or 3D scene datasets~\cite{chang2017matterport3dlearningrgbddata}. In contrast, iterative optimization methods~\cite{chung2023luciddreamerdomainfreegeneration3d, höllein2023text2roomextractingtextured3d, yu2024viewcraftertamingvideodiffusion, zhang2024text2nerftextdriven3dscene, paliwal2024panodreamer} rely on repeated rendering and refinement steps to progressively build up the 3D scene. More recently, warp-and-inpaint approaches have gained attention~\cite{lei2023rgbd2generativescenesynthesis, seo2024genwarpsingleimagenovel, wang2024vistadreamsamplingmultiviewconsistent}. These methods first warp the input view to novel viewpoints using estimated depth or geometry, and then inpaint occluded or missing regions using image-based or feature-based techniques. They often employ intermediate 3D representations to guide refinement and ensure spatial consistency. Our method also follows the warp-and-inpaint paradigm, achieving high appearance consistency and geometric plausibility across the entire scene.

\section{Method}

\begin{figure*}[h]
  \centering
  \includegraphics[width=\textwidth]{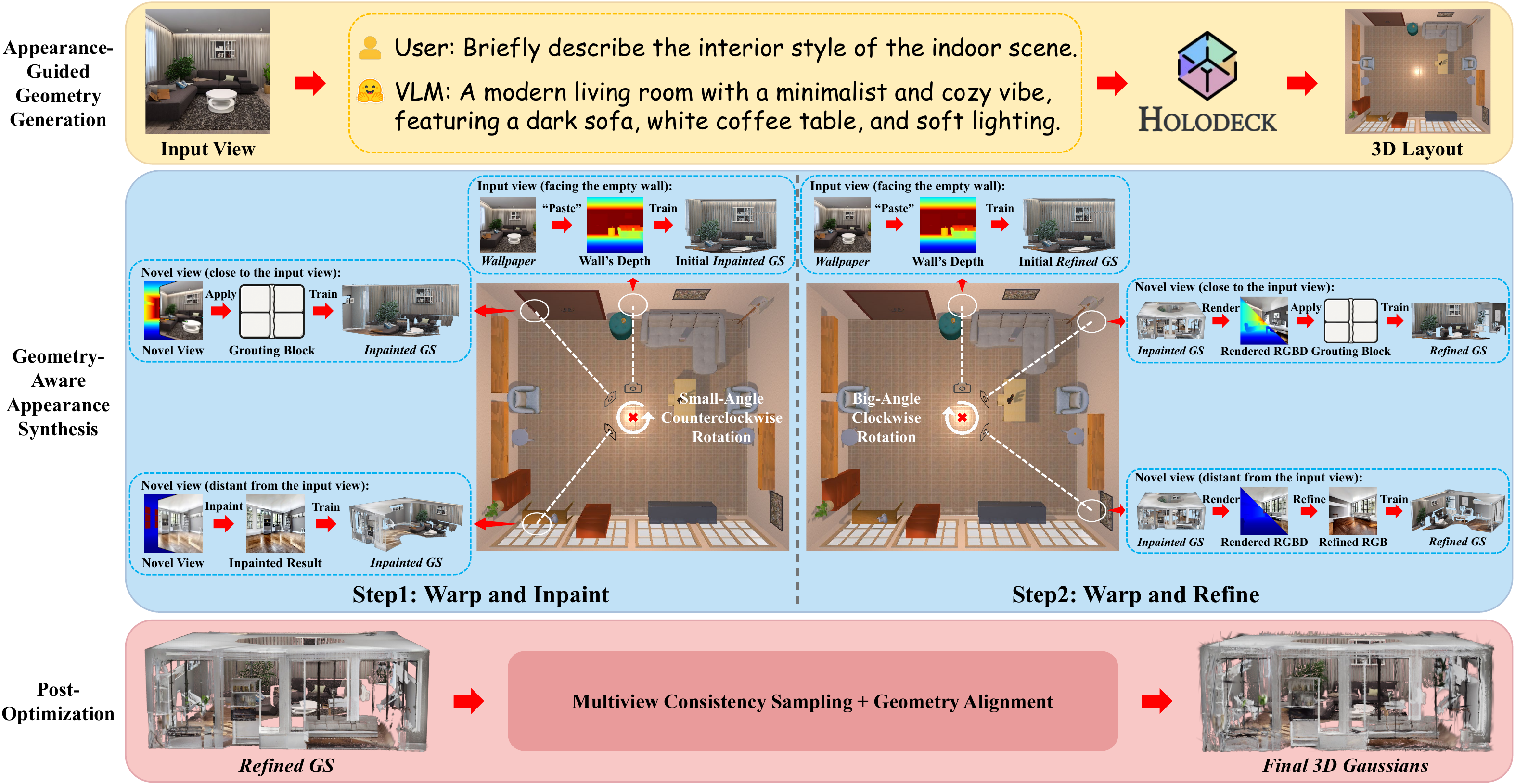}
  \caption{\textbf{Overview of the AnchoredDream pipeline.} Given a single-view image, our method generates a complete 360° indoor scene through appearance-guided geometry generation, geometry-aware appearance synthesis, and post-optimization.}
  \vspace{-25pt}
  \label{fig:pipeline}
\end{figure*}

We propose \textbf{AnchoredDream}, a novel pipeline for generating complete 360° indoor scenes via geometric grounding, as shown in Fig.~\ref{fig:pipeline}. Our method employs an appearance-geometry mutual boosting mechanism to ensure both appearance consistency and geometric plausibility. In the following sections, we first introduce our appearance-guided geometry generation approach (Sec.~\ref{3.1}), followed by the geometry-aware appearance synthesis process (Sec.~\ref{3.2}). Finally, Sec.~\ref{3.3} presents a post-optimization step for refining the 3D scene representation.

\vspace{-15pt}

\subsection{Appearance-Guided Geometry Generation}
\label{3.1}

VLMs have shown strong capabilities in image understanding, capturing both high-level visual attributes and semantic context~\cite{jia2021scalingvisualvisionlanguagerepresentation, li2022blipbootstrappinglanguageimagepretraining, liu2023visualinstructiontuning, radford2021learningtransferablevisualmodels}. Given a single-view input image, our goal is to extract a coarse yet informative prior to guide subsequent layout generation. To this end, we employ a VLM to generate a scene-level description capturing holistic attributes such as visual style (e.g., modern or classical) and scene category (e.g., living room or bedroom). We then feed the description into Holodeck~\cite{yang2024holodecklanguageguidedgeneration} to generate a high-fidelity 3D indoor layout.

During the following appearance synthesis stage, we arbitrarily select one wall from the layout and temporarily project the input image onto it as a wallpaper. The geometry misalignment between the input image and the selected wall is later corrected during post-optimization (Sec.~\ref{3.3.2}). We place a virtual camera at the center of the room, oriented towards the selected wall. As the camera rotates 360°, we render depth maps for each view from the layout to guide the appearance synthesis. We refer to these depth maps as \textit{rendered depth}, which can also be regarded as ground-truth depth.

\subsection{Geometry-Aware Appearance Synthesis}
\label{3.2}

In this section, we first review core diffusion techniques (Sec.~\ref{3.2.1}). We then introduce the \textbf{Grouting Block} (Sec.~\ref{3.2.2}), which seamlessly blends the transition between the input view and the generated regions. Sec.~\ref{3.2.3} presents our warp-and-inpaint strategy, followed by a warp-and-refine mechanism in Sec.~\ref{3.2.4} that further improves visual quality and consistency.

\subsubsection{Controlling Generation via Denoising Strength}
\label{3.2.1}

Recalling the reverse process of an LDM:

\begin{equation}
p_\theta(\mathbf{z}_{t-1} \mid \mathbf{z}_t) = \mathcal{N}(\mathbf{z}_{t-1}; \mu_\theta(\mathbf{z}_t, t), \Sigma_\theta(\mathbf{z}_t, t)),
\end{equation}

where \( t = 1, 2, \dots, T \), \(\mathbf{z}_t\) denotes the latent variable at timestep \( t \), and \(\theta\) represents the parameters of the neural network that predicts the noise in the reverse process.

To control the generation process and balance fidelity to the input with generative flexibility, we introduce a denoising strength parameter \(\gamma\) during the reverse process~\cite{chen2023text2textextdriventexturesynthesis}: \(\gamma = \frac{t}{T}\). Here, \(\gamma\) determines the starting point of the denoising: instead of beginning from pure noise, we start from an intermediate latent \(\mathbf{z}_t\). A lower \(\gamma\) (closer to 0.0) corresponds to starting from a cleaner latent, thus preserving more structure from the input. A higher \(\gamma\) (closer to 1.0) permits greater generative freedom but at the cost of reduced structural alignment. In our method, we leverage \(\gamma\) to control the behavior of the inpainting and refinement stages.

\subsubsection{Grouting Block}
\label{3.2.2}

When we project the input image onto one wall of the layout as a wallpaper, a depth mismatch is introduced because the wall’s geometry differs from the true scene geometry in the image. Consequently, when synthesizing nearby viewpoints, our method renders depth maps consistent with the wall geometry rather than the input image. This discrepancy causes misalignment at the image boundaries, leading to visible inconsistencies in the inpainted regions.

\begin{figure}[h]
  \centering
  \includegraphics[width=\linewidth]{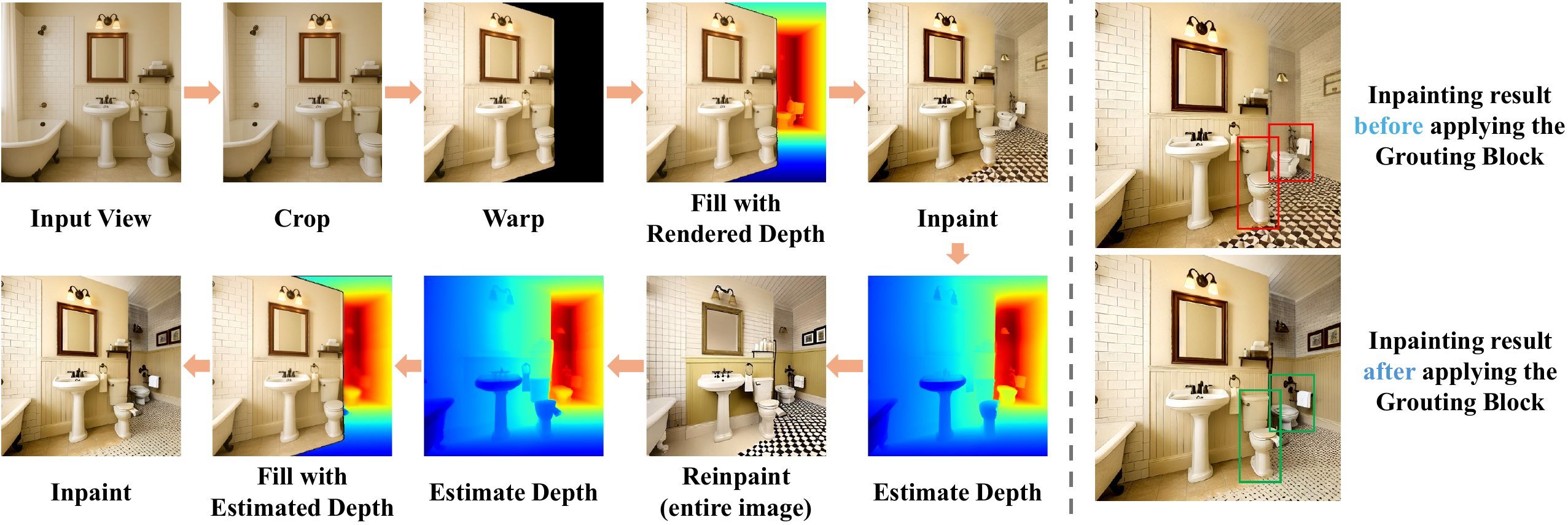}
  \caption{\textbf{Illustration of the Grouting Block.} Our two-stage refinement process ensures seamless blending between the input view and newly generated regions.}
  \vspace{-10pt}
  \label{fig:grouting}
\end{figure}

To address the inconsistencies between the input image and the inpainted regions, enabling smoother transitions, we propose the \textbf{Grouting Block}. The design is inspired by tile grouting in interior renovation, where filler material is used to seamlessly bridge the gaps between tiles and create a smooth, unified surface. It is applied in both the warp-and-inpaint and warp-and-refine stages. In what follows, we describe its use in warp-and-inpaint; its application in warp-and-refine follows a similar design.

We first crop a small margin from both the left and right sides of the input image, removing partial objects near the edges while minimally affecting the overall content. When warping to a novel viewpoint near the input view, we first render the corresponding depth map from the layout and inpaint the unseen regions under this depth guidance. We then apply a MDE model to the inpainted image to predict its depth. However, due to discontinuities around transition regions, the estimated depth may still exhibit local inconsistencies. Nevertheless, thanks to the smoothing capability of MDE models, these inconsistencies are significantly reduced compared to directly stitching the rendered depth. We then use this estimated depth to guide a depth-conditioned diffusion model, with \(\gamma\) set close to 1.0, to inpaint the entire image rather than only the initially unknown regions. Leveraging the generative capabilities of diffusion models, this process effectively smooths artifacts at the boundary between the input image and the inpainted regions. The depth guidance also helps preserve the overall spatial structure, avoiding unintended shifts in object layout. Finally, we reapply depth estimation on the reinpainted image to obtain a cleaner depth map. This refined depth is then used to complete the initial inpainting, restricted to the originally unknown regions. A more intuitive way to understand our Grouting Block is to refer to Fig.~\ref{fig:grouting}.

\subsubsection{Warp and Inpaint}
\label{3.2.3}

We begin by using a VLM to extract key scene attributes (distinct from the scene-level description in Sec.~\ref{3.1}) from the single-view input image, including scene category, interior style, wall color, and floor color. These attributes are then composed into a structured textual prompt, referred to as the \textit{scene core}. 

We place a virtual camera at the center of the layout, oriented toward the selected wall. Starting from the input view, we perform warp-and-inpaint operations in a counterclockwise direction.

At each step, the inpainted image is appended to a frame list, which is used to incrementally train a 3D Gaussian representation of the scene, referred to as the \textit{inpainted GS}. When warping to a novel viewpoint, we render the corresponding image from the current \textit{inpainted GS} and use the rendered alpha map to identify unknown regions that require inpainting. 

If the novel view is close to the input view, we apply our proposed Grouting Block to perform inpainting. For more distant views, we render a depth map from the layout and feed it, along with the \textit{scene core}, into a depth-conditioned diffusion model to guide inpainting of the unknown regions. We set \(\gamma\) to 1.0 to enable fully generative synthesis.

To ensure continuity and maintain appearance consistency, each warp step only spans a small angular increment. This keeps the unknown region compact, allowing the inpainting model to leverage nearby visible regions for effective appearance propagation.

\subsubsection{Warp and Refine}
\label{3.2.4}

After completing the warp-and-inpaint stage, we obtain a coarse 3D Gaussian representation of the scene, which we refer to as the \textit{inpainted GS}. In the subsequent warp-and-refine stage, we gradually construct a new 3D Gaussian representation, denoted as the \textit{refined GS}. 

This time, we traverse viewpoints in a clockwise direction, in contrast to the earlier counterclockwise sweep. Our intuition is that inpainting in the counterclockwise direction primarily leverages appearance information from the left side of the input image; thus, traversing in the opposite direction helps enhance consistency using the right-side appearance cues. 

For each new viewpoint, we adopt different strategies based on its distance to the input view. If the viewpoint is close to the input, we apply our Grouting Block to perform localized refinement. For more distant views, we first render an RGB image from the \textit{inpainted GS} and obtain the rendered depth map from the layout. We then use a VLM to generate a textual prompt of the rendered RGB image. The unknown region mask is derived from the on-the-fly trained \textit{refined GS}. With these inputs, we use a depth-conditioned diffusion model to refine the rendered RGB image, applying a moderate \(\gamma\) (close to 0.5). This strategy helps to preserve previously synthesized appearance information while correcting artifacts and inconsistencies in the transition regions.

Each warp step spans a relatively large angle to maximize refinement efficiency. Since a coarse foundation has already been established by the \textit{inpainted GS}, a larger unknown region does not significantly impact the quality of refinement. Moreover, larger viewpoint transitions help mitigate inconsistencies arising from objects that appear partially in one view and fully in the next, which would otherwise lead to inconsistent appearances of the same object.

\subsection{Post-Optimization}
\label{3.3}

\subsubsection{Appearance}

After completing the previous stages, we obtain a refined 3D Gaussian representation, referred to as the \textit{refined GS}. However, this representation may still exhibit slight multi-view inconsistencies. To address this, we adopt Multi-view Consistency Sampling~\cite{wang2024vistadreamsamplingmultiviewconsistent} to further enhance visual coherence across views. Recalling the reverse denoising process in diffusion models, and excluding the encoder and decoder for simplicity, a single denoising step can be expressed as:

\begin{equation}
\label{con:1}
\mathbf{x}_{t-1} = s_t \mathbf{x}_t + d_t \mu_t + \sigma_t \epsilon,
\end{equation}
where \(\mu_t\) is the estimation of the final denoising result \(\mathbf{x}_0\), and \(s_t\), \(d_t\), and \(\sigma_t\) are predefined constants. \(\epsilon\) is standard Gaussian noise. This formulation indicates that we can impose constraints on each denoising step by modifying \(\mu_t\): given \(N\) viewpoints, at every denoising step, we first use the \textit{refined GS} to render the corresponding images \(\overline\mu_t^{1:N}\) which serve as multi-view references, and then we adjust the original \(\mu_t\) using the rendered guidance:

\begin{equation}
\hat{\mu}_t = w_t \cdot \phi_t \overline{\mu}_t + (1 - w_t) \cdot \mu_t,
\end{equation}
where \(\phi_t = \frac{std(\mu_t)}{std(\overline\mu_t)}\) normalizes the dynamic range of the two signals to avoid overexposure~\cite{lin2024commondiffusionnoiseschedules}, and \(w_t\) is a hyperparameter balancing smoothness and detail preservation. At each denoising step, we substitute \(\hat\mu_t\) for \(\mu_t\) in Eq.~\ref{con:1}, thereby imposing the multi-view consistency constraint. Simultaneously, the rectified signals \(\hat\mu_t^{1:N}\) are used to update the \textit{refined GS}, ultimately producing a more visually consistent 3D representation.

\subsubsection{Geometry}
\label{3.3.2}

\vspace{-30pt}
\begin{figure}[ht]
  \centering
  \includegraphics[width=\linewidth]{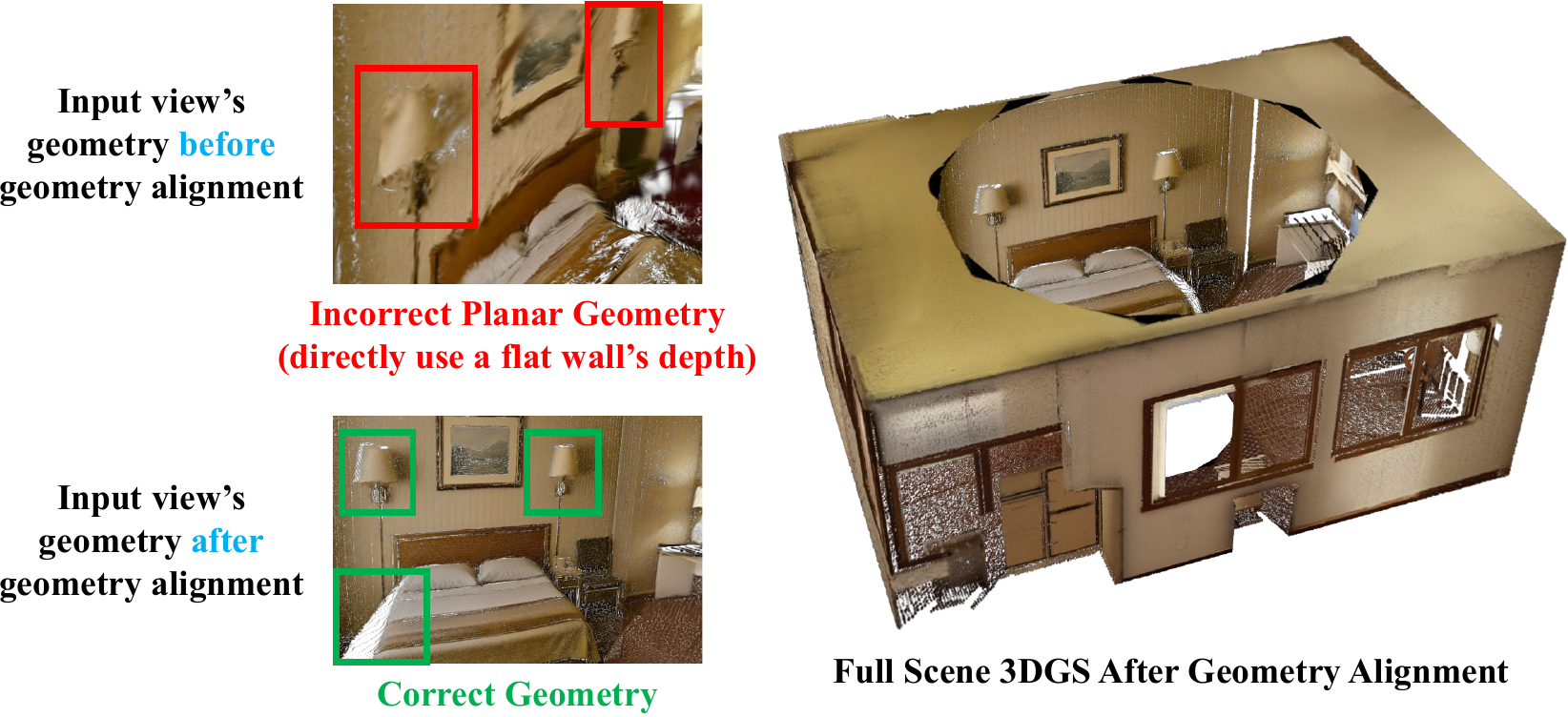}
  \caption{\textbf{Effect of geometry alignment on scene geometry (case: ``A vintage-style budget hotel room'').} Geometry alignment effectively rectifies input view's geometry errors while maintaining the global geometry consistency of the full scene.}
  \vspace{-10pt}
  \label{fig:geometry_alignment}
\end{figure}

We adopt a simple yet effective approach to address the geometry misalignment between the input image and the selected wall (Fig.~\ref{fig:geometry_alignment}). Specifically, we first use an MDE model to predict the depth of the input image, denoted as $\mathbf{\hat{D}}$. Given the rendered depth of the selected wall, $\mathbf{D}$, we then estimate the optimal affine transformation parameters:

\begin{equation}
\alpha^*, \beta^* = \underset{\alpha, \beta}{\arg\min}
\sum_{(u,v)}
\left\|
\left( \frac{\alpha}{\mathbf{\hat{D}}} + \beta - \frac{1}{\mathbf{D}} \right)
\right\|_2^2,
\end{equation}
and the final aligned depth map is computed as $\mathbf{D^*} = \frac{1}{\alpha^* / \mathbf{\hat{D}} + \beta^*}$. During 3DGS optimization, we replace the selected wall's depth with this aligned depth map, while retaining the rendered depth for all other views.

\section{Experiments}

\begin{figure*}[t]
  \centering
  \includegraphics[width=\textwidth]{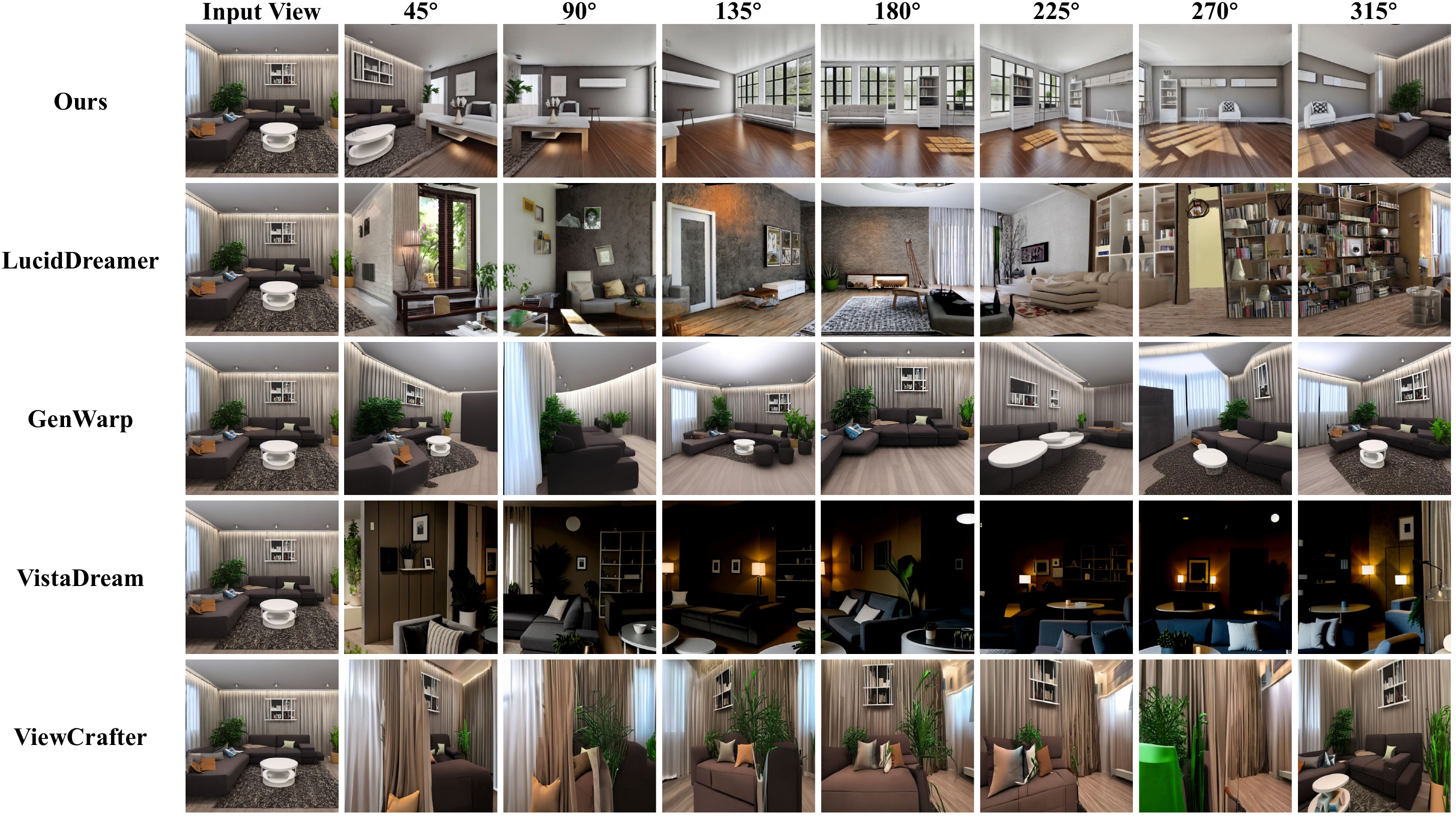}
  \caption{\textbf{Visual quality comparison for the case “A modern-style living room”.} Our method faithfully preserves the appearance style of the input image, whereas the baselines exhibit visual artifacts, style drift, and inconsistencies in the generated results.}
  \vspace{-10pt}
  \label{fig:visual quality}
\end{figure*}

\subsection{Experimental Setup}
\label{4.1}

\paragraph{Datasets}
We evaluate our method on a curated set of 25 single-view indoor images, each representing a distinct and commonly encountered scene category (e.g., arcade, bedroom, gym, living room). The dataset also covers a diverse range of interior design styles, including Retro, Modern, Minimalist, and Modern Luxury. Among these images, 23 are generated using GPT-4o~\cite{openai2024gpt4ocard}, while the remaining 2 are adopted from prior baseline works, as these images are widely used in existing evaluations.

\paragraph{Baselines}
We evaluate our approach against several representative baselines for single-view scene generation, including LucidDreamer~\cite{chung2023luciddreamerdomainfreegeneration3d}, GenWarp~\cite{seo2024genwarpsingleimagenovel}, VistaDream~\cite{wang2024vistadreamsamplingmultiviewconsistent}, and ViewCrafter~\cite{yu2024viewcraftertamingvideodiffusion}. These methods are chosen for their demonstrated capability in novel view synthesis and their potential to generate complete 360° indoor scenes. We exclude RealmDreamer~\cite{shriram2025realmdreamertextdriven3dscene} because its codebase does not currently support custom 360° view trajectories. Moreover, as reported in its original paper, LucidDreamer achieves comparable performance; thus, we adopt it as a substitute to ensure a fair evaluation.

\subsection{Qualitative Results}

\paragraph{Visual Quality}
Fig.~\ref{fig:visual quality} shows the input image and the corresponding synthesized views at 45°, 90°, 135°, 180°, 225°, 270° and 315° generated by our method and four baselines. \textbf{LucidDreamer} exhibits significant inconsistency in appearance across viewpoints. The synthesized images often display abrupt style changes, suggesting that visual cues from the input image are not effectively preserved during generation. \textbf{GenWarp} and \textbf{ViewCrafter} suffer from severe visual artifacts, including ghosting, blurry textures, and structural distortions. Notably, their results at 180° appear overly similar to the input image, indicating that the models may be rotating the input view rather than synthesizing novel content aligned with a fixed-camera 360° rotation. \textbf{VistaDream} often generates scenes that are noticeably darker and underexposed compared to the input. This leads to a visible loss of fine-grained details. More critically, it sometimes produces sudden geometric anomalies, such as unexpected corners or entire room segments, indicating a lack of spatial consistency across views. {In contrast, \textbf{our method} maintains the overall style and fine-grained appearance of the input view across all angles.

\paragraph{Geometric Plausibility}
Fig.~\ref{fig:geometric plausibility} compares the reconstructed 3DGS produced by our method and three baseline methods (excluding \textbf{GenWarp}, which does not produce point clouds). Both \textbf{LucidDreamer} and \textbf{VistaDream} generate enclosed scene structures. However, the resulting scenes often exhibit irregular shapes, unexpected corners, and deviate from the typical rectangular layouts commonly found in indoor environments. Another common failure case is \textbf{ViewCrafter}, which produces highly incomplete reconstructions. Its point clouds resemble direct reprojections of the input view, with minimal scene expansion. This explains why its synthesized images at 180° appear nearly identical to the input image. \textbf{Our method}, by comparison, generates a well-formed rectangular cuboid structure that closely aligns with the expected geometry of indoor scenes.

\begin{figure*}[t]
  \centering
  \includegraphics[width=\textwidth]{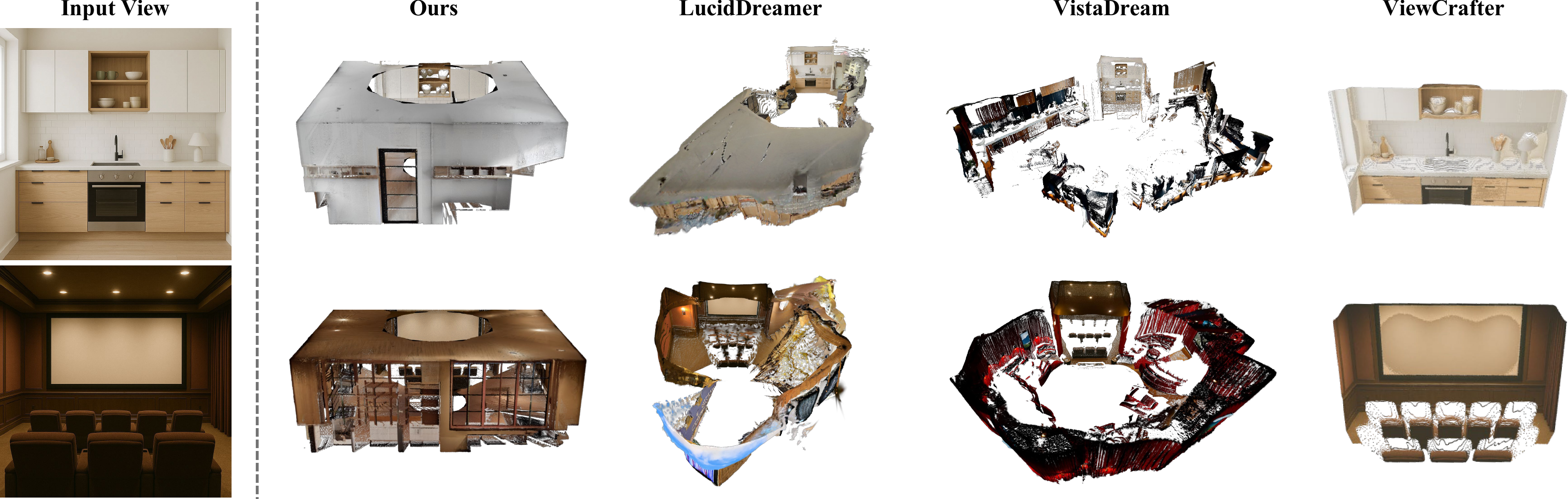}
  \caption{\textbf{Geometric plausibility comparison for the cases “A Scandinavian-style
    kitchen” (top) and “A modern-style home theater” (bottom).} Our method generates well-structured, rectangular scenes, whereas the baselines produce irregular or incomplete geometries.}
    \vspace{-10pt}
  \label{fig:geometric plausibility}
\end{figure*}

\subsection{Quantitative Results}

\begin{table}[ht]
\centering
\caption{
\textbf{Quantitative comparison of appearance consistency and geometric plausibility, evaluated using LLaVA-NeXT.} Our method consistently achieves the highest scores on both metrics, significantly outperforming LucidDreamer, and VistaDream.
}
\label{tab:quantitative}
\resizebox{\linewidth}{!}
{
\begin{tabular}{lccc}
\toprule
Metric & LucidDreamer & VistaDream & Ours \\
\midrule
Appearance Consistency~\(\uparrow\) & 0.943 & 0.477 & \textbf{0.957} \\
Geometric Plausibility~\(\uparrow\) & 0.286 & 0.651 & \textbf{0.808} \\
\bottomrule
\end{tabular}
}
\vspace{-15pt}
\end{table}

We conduct a quantitative comparison on two key metrics: \textbf{Appearance Consistency} and \textbf{Geometric Plausibility}. Our method is evaluated against LucidDreamer and VistaDream, as the other baselines do not support full 360° scene generation and instead rely on rotating the input view without synthesizing novel content. For each method, we render views along the same camera trajectory and evaluate the results using LLaVA-NeXT~\cite{liu2024llavanext}. To assess appearance consistency, we evaluate whether the rendered views remain coherent with the input in terms of visual style, lighting, and dominant color. For geometric plausibility, we consider the plausibility of spatial layout, object proportions, and overall structural coherence. As shown in Tab.~\ref{tab:quantitative}, our method significantly outperforms all baseline approaches on both evaluation metrics, clearly demonstrating its effectiveness and robustness in generating visually consistent and geometrically sound 360° indoor scenes.

\subsection{Ablation Study}

\vspace{-20pt}
\begin{figure}[ht]
  \centering
  \includegraphics[width=\linewidth]{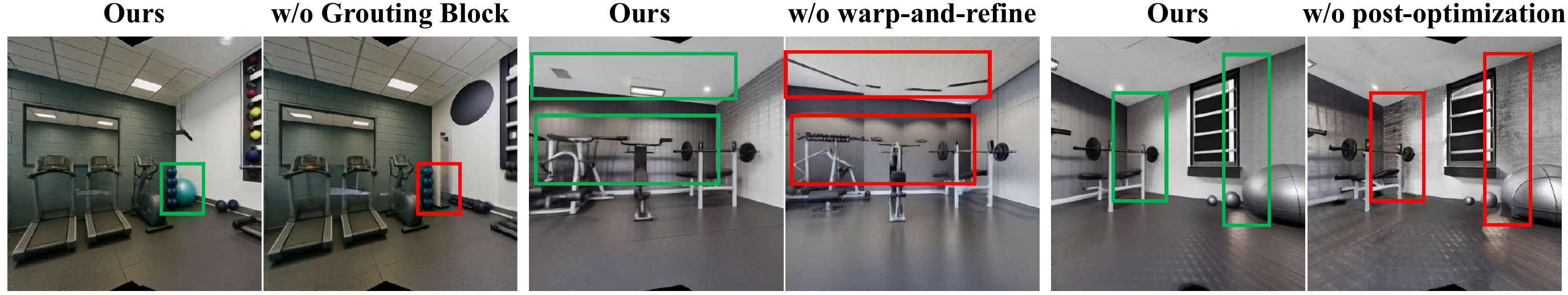}
  \caption{\textbf{Ablation study results.} Removing individual components leads to noticeable artifacts and reduced appearance consistency, highlighting the importance of each module in our pipeline.}
  \vspace{-10pt}
  \label{fig:ablation}
\end{figure}

We remove key components in our pipeline to assess their individual contributions, as illustrated in Fig.~\ref{fig:ablation}. Removing the \textbf{Grouting Block} results in noticeable appearance inconsistencies around the input view, introducing strong disconnection artifacts and visual discontinuities in the transition regions. Removing the \textbf{warp-and-refine} stage leads to duplicated or fragmented object appearances—e.g., a single object may exhibit inconsistent colors or textures even within the same viewpoint due to uncorrected inpainting artifacts—and introduces visible seam artifacts between adjacent frames. Finally, removing the \textbf{post-optimization} (appearance) step reduces overall appearance smoothness, causing unnatural transitions in certain regions. In summary, each component of our method plays a critical role in maintaining appearance consistency and ensuring high-quality 360° scene generation.

\subsection{Discussion on Panorama-Based Methods}

Our method addresses the problem of 360° full indoor scene generation, and recent panorama-based methods~\cite{chen2025splatter, huang2025scene4u, yang2025layerpano3d, zhou2024dreamscene360} have also demonstrated their potential for this task. However, our setting follows an image-to-scene paradigm, taking a single-view image as input to generate the complete scene. In contrast, panorama-based methods are typically developed under a text-to-scene setting, using text prompts as input. Nevertheless, a brief comparison with these methods is still worthwhile, and we include DreamScene360~\cite{zhou2024dreamscene360} as a representative baseline. Fig.~\ref{fig:panorama} presents the visual results.

\vspace{-20pt}
\begin{figure}[ht]
  \centering
  \includegraphics[width=\linewidth]{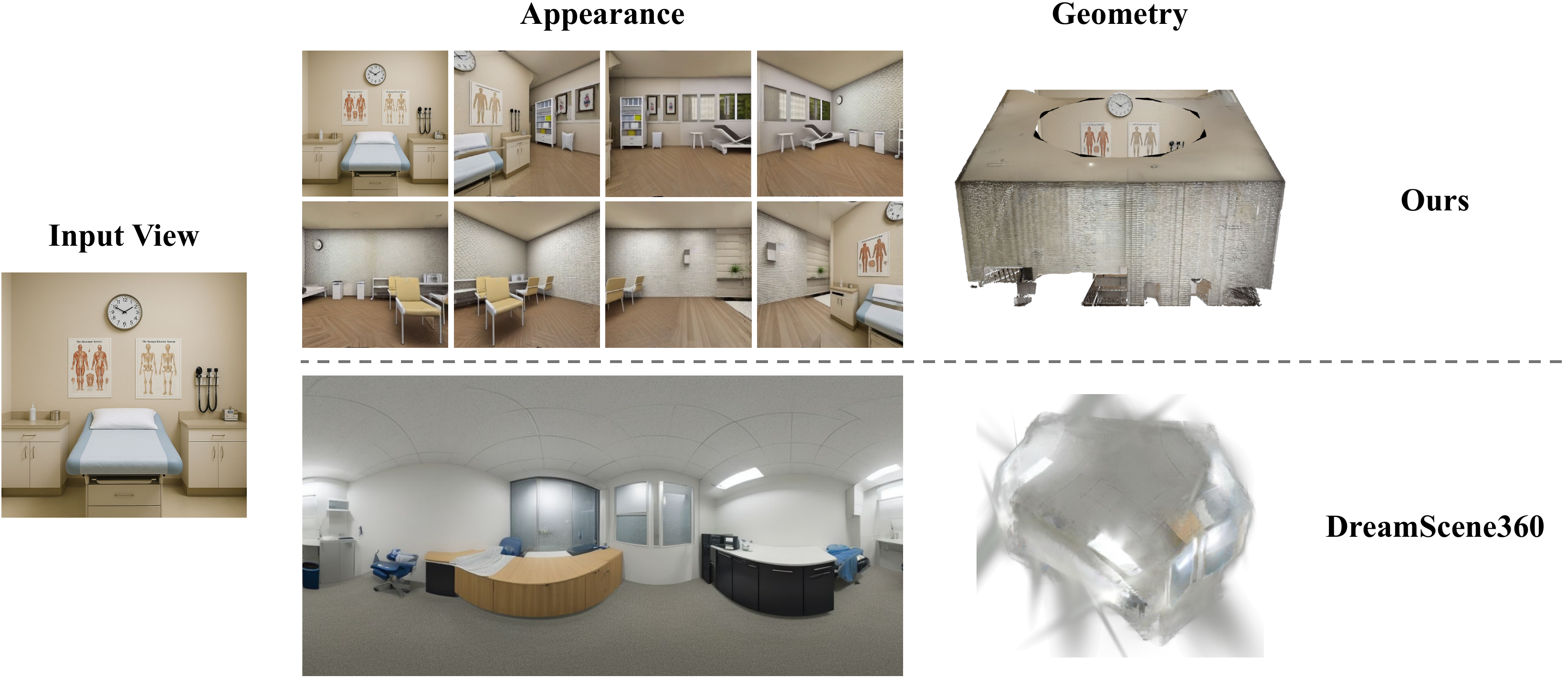}
  \caption{\textbf{Appearance and geometry comparison with DreamScene360 for the case “A sterile modern-style medical examination room”.} Our method generates scenes that preserve the appearance of the input view and exhibit rectangular room geometries. In contrast, DreamScene360 shows a substantial appearance mismatch with the input view and produces bulging, spherical-like structures.}
  \vspace{-25pt}
  \label{fig:panorama}
\end{figure}

\paragraph{Appearance Consistency}

DreamScene360 often struggles to seamlessly integrate a single-view input image into the final output. Under its text-to-scene paradigm, we first describe the input view and then use this textual description to guide panorama generation, followed by scene reconstruction. However, textual descriptions inevitably omit fine-grained visual details, resulting in appearance inconsistencies. In contrast, our method directly leverages the original single-view image, enabling better preservation of appearance and more seamless fusion into the generated scene.

\paragraph{Geometric Plausibility}

With DreamScene360, the generated panoramas typically exhibit reasonably plausible layouts, such as rooms with four corners and minimal distortion, but the corresponding 3D point clouds often suffer from poor geometry. In particular, they tend to bulge outward, with corners nearly disappearing, resulting in a spherical-like structure. By comparison, our method produces final results with more realistic and structurally coherent room geometries.

\subsection{Application: Text-to-Scene Generation}
\vspace{-5pt}

\begin{figure}[ht]
  \centering
  \includegraphics[width=\linewidth]{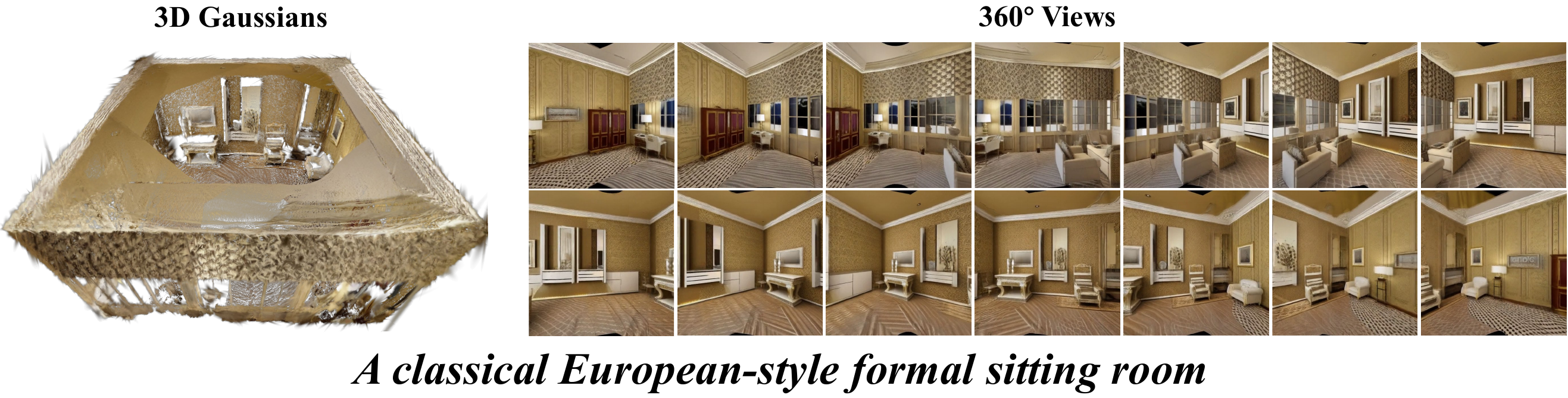}
  \vspace{-15pt}
  \caption{\textbf{Text-to-scene generation.} Given an indoor scene description, our method generates a visually consistent and geometrically plausible 360° scene.}
  \vspace{-10pt}
  \label{fig:text2scene}
\end{figure}

Our method can be naturally extended to text-to-scene generation, as illustrated in Fig.~\ref{fig:text2scene}. Given an indoor scene description, we first generate a 3D layout and then execute our pipeline using only the textual description and the layout, without any input view. Results show that our method continues to achieve high appearance consistency and geometric plausibility under this purely text-driven setting.

\section{Limitations}
\vspace{-5pt}

While AnchoredDream demonstrates strong performance in single-view 360° indoor scene generation, achieving high appearance consistency and geometric plausibility, it still has a few limitations. First, the method relies on prompts generated by VLMs, and biased or inaccurate scene descriptions may negatively affect generation quality. However, this issue is expected to diminish as VLMs continue to improve in scene understanding and prompt reliability. Second, our pipeline employs ControlNet~\cite{zhang2023addingconditionalcontroltexttoimage} for depth-guided inpainting, as it remains the most effective and practical solution available to date. However, it occasionally fails to propagate appearance information from known to unknown regions, leading to local inconsistencies. This issue could be alleviated in future work by developing more powerful depth-guided inpainting models or by leveraging advanced image editing models such as Nano Banana for appearance refinement.

\section{Conclusion}
\vspace{-5pt}

We have presented AnchoredDream, a novel pipeline for single-view indoor scene generation grounded on high-fidelity geometry. By introducing an appearance-geometry mutual boosting mechanism, our method can generate complete 360° indoor scenes from a single image while enhancing both visual and geometric consistency. With the proposed Grouting Block, we further achieve seamless blending between the input view and the generated regions. Extensive experiments demonstrate that AnchoredDream outperforms existing baselines in both appearance consistency and geometric plausibility, while operating in a zero-shot manner. We believe AnchoredDream offers a promising direction for scalable, geometry-aware scene generation and holds strong potential for real-world applications such as robotics and virtual content creation.

\bibliographystyle{splncs04}
\bibliography{main}

\clearpage
\appendix

\section{Implementation Details}

\subsection{Computing Resources}

All experiments were conducted on a single NVIDIA GeForce RTX 3090 GPU with 24 GB of memory, running Ubuntu 20.04 and CUDA 11.8 for GPU acceleration.

\subsection{Pretrained Models}

We consistently use LLaVA~\cite{liu2023visualinstructiontuning} as our VLM, Holodeck~\cite{yang2024holodecklanguageguidedgeneration} as our 3D layout generator, ControlNet~\cite{zhang2023addingconditionalcontroltexttoimage} as our depth-guided inpainting model, and \mbox{Depth Pro~\cite{bochkovskii2025depthprosharpmonocular}} as our MDE model.

The primary reason we chose LLaVA is that it is a widely used open-source VLM, freely available and demonstrating strong overall performance. In most VLM use cases within our pipeline, such as identifying room category, interior style, or overall ambiance, highly precise descriptions are not required, and LLaVA sufficiently meets our needs, even though other models (e.g., GPT-4o~\cite{openai2024gpt4ocard}, Gemini~\cite{gemini2025}) may offer better performance. However, during the warp-and-inpaint stage, more precise and detailed color descriptions can significantly improve the final results. Therefore, we recommend users to opt for GPT-4o in the warp-and-inpaint stage when possible. Nonetheless, since our project is open-sourced and we aim to make it as accessible as possible to a broad range of users, we have chosen LLaVA as the default VLM.

\subsection{Stage-Wise Viewpoint Configuration}

In the warp-and-inpaint stage, we use 20 novel viewpoints; in the warp-and-refine stage, we use 8 novel viewpoints; and in the post-optimization (multi-view consistency sampling) stage, we use 15 viewpoints.

\subsection{Viewpoint Distance Definition}

In the warp-and-inpaint stage, we define close-view angles as those within (0°, 90°) $\cup$ (270°, 360°), and the distant views as the remaining angles. In the warp-and-refine stage, the camera rotates with larger angular intervals between viewpoints, and we define the first and last viewpoints as close views, and all others as distant views, since by the second and second-to-last viewpoints, the input image is no longer visible.

\subsection{Usage of the Grouting Block}

The Grouting Block is applied only to close views, and not to distant ones. Its main purpose is to seamlessly blend the transition between the input view and newly generated regions. For distant views, where the input image is no longer visible, the Grouting Block is unnecessary, as the geometry of the generated layout transitions naturally and smoothly, without introducing any noticeable artifacts or unnatural boundaries between views.

\subsection{3DGS Optimization}

All 3DGS representations used in our method are based on original Gaussians~\cite{kerbl20233dgaussiansplattingrealtime}, rather than feature Gaussians.

The 3DGS instances described in Sec.~\ref{3.2.3} and Sec.~\ref{3.2.4} are reconstructed independently. In Sec.~\ref{3.2.3}, we reconstruct a 3DGS instance referred to as the \textit{inpainted GS}. In Sec.~\ref{3.2.4}, we reconstruct another instance called the \textit{refined GS}. Specifically, the \textit{inpainted GS} serves as a coarse scaffold, and in Sec.~\ref{3.2.4}, we refine the appearance of each view based on the rendered images from the same viewpoints in the \textit{inpainted GS}. These refined images are then used to reconstruct a new 3DGS instance—the \textit{refined GS}.

Different from conventional 3DGS optimization methods that initialize Gaussian primitives from a point cloud and update their parameters using supervision from all viewpoints, our method adopts an incremental optimization paradigm. At the beginning (when the camera faces the input view), we initialize Gaussian primitives using the reprojected point map of the input image. The camera intrinsics and extrinsics are known, since they are shared with the layout depth rendering process. We then train the Gaussians under the supervision of the input image. As the camera rotates and a new inpainted image becomes available, we reproject it using the corresponding rendered depth to initialize new Gaussians for the unseen regions. Subsequently, the 3DGS instance is trained under the supervision of all images accumulated up to that point. We provide pseudocode of the 3DGS optimization process during the warp-and-inpaint stage for clarity (see Algorithm~\ref{alg:anchoreddream}).

\begin{algorithm}[ht]
\caption{3DGS Optimization}
\label{alg:anchoreddream}
\KwIn{Input view $I_\text{in}$, generated layout $L$, camera rotation sequence $\{v_i\}$}
\KwOut{Optimized 3DGS}

\textbf{Initialize:} \\
\Indp
input\_mask $\leftarrow$ all\_ones\_mask(same\_size\_as($I_\text{in}$))\;
frame\_list $\leftarrow$ [($I_\text{in}$, input\_mask)]\;
init\_viewpoint $\leftarrow$ viewpoint facing the wall\;
init\_depth $\leftarrow$ render\_depth(init\_viewpoint, $L$)\;
3DGS $\leftarrow$ train\_3DGS($I_\text{in}$, init\_depth)\;
\Indm

\For{each $v$ in camera\_rotation\_sequence}{
  rendered\_image, mask $\leftarrow$ render\_rgb($v$, 3DGS)\;
  rendered\_depth $\leftarrow$ render\_depth($v$, $L$)\;
  inpainted\_image $\leftarrow$ inpaint(rendered\_image, rendered\_depth, mask)\;
  frame\_list.append((inpainted\_image, mask))\;

  \textbf{Optimize 3DGS:}\\
  \Indp
  \For{many iterations}{
    (image, mask) $\leftarrow$ randomly sample from frame\_list\;
    valid\_pixels $\leftarrow$ apply(mask, image)\;
    update 3DGS using valid\_pixels\;
  }
  \Indm
}

\Return{Optimized 3DGS}
\end{algorithm}

It is true that directly using images generated by diffusion models can introduce random noise. However, we have implemented several mechanisms to address this issue and ensure robust 3DGS reconstruction:
We do not use diffusion models in a purely text-driven manner. Instead, we employ depth-guided diffusion models (specifically, ControlNet~\cite{zhang2023addingconditionalcontroltexttoimage}) to ensure reliable image synthesis. We first render reliable depth maps, which already contain deterministic scene structure and object layout. Thanks to ControlNet’s strong guidance ability, the synthesized appearance corresponds accurately to these depth maps, avoiding the issues of hallucinated or missing objects. Also, we incorporate multiple stages in our pipeline, including the Grouting Block, warp-and-inpaint, warp-and-refine, and post-optimization, to iteratively refine scene appearance and maintain global consistency. At each stage, we progressively enhance the 3DGS to suppress random noise such as color mismatches. Our final results demonstrate that we achieve a consistent scene appearance, both in terms of style and dominant color.

\subsection{Quantitative Metrics}

We adopt LLaVA-NeXT~\cite{liu2024llavanext} as our VLM instead of LLaVA~\cite{liu2023visualinstructiontuning}, since it supports multi-image inputs, which better aligns with our experimental setup.

We use each method’s 360° rendering video as input and uniformly sample 15 frames, which we found sufficient to cover the room’s visual content. Since some baselines produce shorter videos, sampling more frames would cause inconsistencies. Thus, 15 frames is a balanced and fair choice across all methods.

For both metrics (Appearance Consistency and Geometric Plausibility), we compute the proportion of "Yes" responses across the 15 sampled frames as the evaluation score for that scene. We perform this evaluation across all scenes in our dataset and report the mean score as the final quantitative result.

\paragraph{Appearance Consistency}

We employ LLaVA-NeXT to evaluate whether the appearance of each sampled frame is consistent with the input view. The evaluation process consists of two steps:

\begin{itemize}
    \item First, we prompt LLaVA-NeXT to extract the input view’s appearance features using the following instruction: ``Describe the image's visual style, lighting, and dominant color using one word for each. List them in this order, separated by commas. For example: `modern, bright, beige'."
    \item Then, we prompt LLaVA-NeXT to judge whether a sampled frame is consistent with the reference using: ``The reference style is as follows: Visual style: xxx, Lighting: xxx, Dominant color: xxx. Based on the image, are the visual style, lighting and dominant color consistent with the reference? Respond with a single word: Yes or No."
\end{itemize}

\paragraph{Geometric Plausibility}

For geometry evaluation, we directly ask LLaVA-NeXT to assess whether each sampled frame appears geometrically plausible, without reference to the input view. The prompt used is: ``Does the scene in this image appear geometrically plausible, with a realistic spatial layout, natural proportions, and coherent 3D structure? Pay special attention to sudden or unnatural corners, broken geometry, or inconsistent depth transitions. Respond only with a single word: Yes or No."

\begin{figure}[ht]
  \centering
  \includegraphics[width=0.85\linewidth]{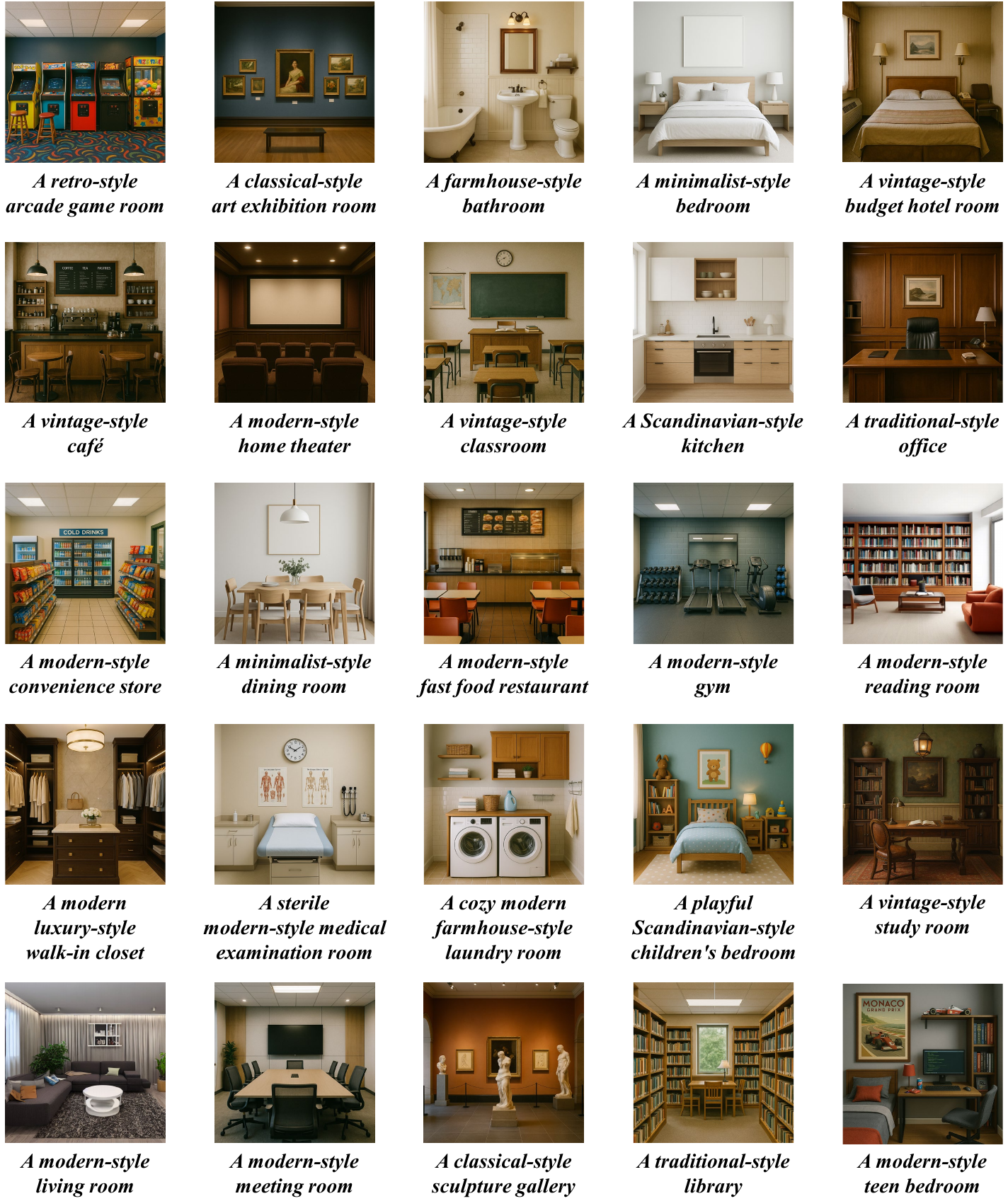}
  \caption{\textbf{Overview of the dataset used in our experiments.} It includes 25 single-view indoor images with diverse scenes and styles, all captured from a frontal wall perspective.}
  \label{fig:dataset}
\end{figure}

\section{Datasets}

We illustrate the dataset used in our experiments in Fig.~\ref{fig:dataset}, which consists of 25 single-view images capturing diverse indoor scenes and styles. Two images (a reading room and a living room) are adopted from prior works, while the remaining 23 are generated using GPT-4o~\cite{openai2024gpt4ocard}. All images depict a frontal view of a single wall, rather than a corner or overhead perspective.

Our decision to present the input image as captured from a viewpoint perpendicular to a wall was made primarily for simplicity and clarity in conveying the core idea. However, our method is not constrained to this specific setting. For example, if the input view faces a corner, our pipeline can still accommodate such configurations. In these cases, during the Geometry-Aware Appearance Synthesis stage, we can adjust the camera pose to align the viewing angle of the generated layout with that of the input view. While this may require some additional engineering effort, the underlying pipeline remains fully applicable without any fundamental changes. We also plan to explore automated camera pose alignment in future work. In this paper, however, we deliberately focused on the simplest and most intuitive setting to better communicate the core idea without introducing unnecessary complexity.

\section{Discussion on Scene Shape}

Although we demonstrate rectangular generated scenes, our pipeline is capable of producing non-rectangular layouts. Based on our experiments, Holodeck~\cite{yang2024holodecklanguageguidedgeneration} can generate polygonal room structures but struggles with curved shapes, such as circles or ellipses. In real-world scenarios, most indoor rooms follow polygonal layouts, with only a small fraction—e.g., the Oval Office—featuring irregular or curved shapes. Therefore, our method remains applicable to the vast majority of real-world settings.

\section{Impact of VLM Errors}

We next discuss how VLM errors can impact the final generated results. However, by employing carefully designed prompts, we are able to eliminate the following types of VLM errors, thereby ensuring high-quality generation.

\subsection{Appearance-Guided Geometry Generation}

In this stage, the goal is to generate a scene-level description capturing holistic attributes, with particular focus on semantic categories, interior design style.

If the description is inaccurate, the generated 3D layout may suffer from semantic errors and stylistic inconsistencies. These issues can result in significant mismatches in the final output, for example, a living room might be misclassified as a kitchen, a Nordic style might be rendered as vintage, and the objects in the generated layout might not align with those in the input view (e.g., a mirror frame with curling ornamental carvings might be appropriate in a vintage-style room but would be highly \noindent\mbox{unlikely in a Nordic one).}

\subsection{Warp-and-Inpaint}

In this stage, the VLM is responsible for describing room-level attributes, including the room category, design style, wall color, and floor color. These collectively constitute the \textit{scene core}. Inaccuracies in these attributes can lead to visual inconsistencies during the ControlNet~\cite{zhang2023addingconditionalcontroltexttoimage} inpainting process. In particular, the precision of color descriptions has a direct impact on visual fidelity, for instance, accurately distinguishing between “blue” and a “cooler, cement-toned industrial bluish-gray” can significantly \noindent\mbox{affect the perceived appearance.}

\subsection{Warp-and-Refine}

In this stage, to maintain consistent appearance throughout the scene, prompts given to the VLM should explicitly include phrases like “ensure style consistency”. Without such instructions, errors may compound over successive iterations, potentially resulting in issues such as the same object appearing in different colors or the same room exhibiting multiple, conflicting dominant color tones.

\section{Runtime Analysis}

\subsection{Module-Wise Runtime Breakdown}

\begin{table}[ht]
\centering
\caption{Runtime breakdown by module.}
\label{tab:module_runtime}
\begin{tabular}{lccc}
\toprule
 Module & Runtime (min) \\
\midrule
Appearance-Guided Geometry Generation & 3.22 \\
Warp and Inpaint & 7.15 \\
Warp and Refine & 4.64 \\
Post-Optimization & 14.95 \\
\bottomrule
\end{tabular}
\end{table}

We provide a module-wise runtime breakdown in Tab.~\ref{tab:module_runtime}. The appearance-guided geometry generation takes approximately 3 minutes, primarily due to the latency of the LLM when generating the scene layout via Holodeck~\cite{yang2024holodecklanguageguidedgeneration}. The warp-and-inpaint stage requires about 7 minutes to process a camera trajectory consisting of 20 viewpoints. Warp-and-refine takes around 5 minutes for an 8-view trajectory. Finally, post-optimization takes approximately 15 minutes to enforce appearance consistency across 15 rendered views.

\subsection{Runtime Comparison with Baselines}

\begin{table}[ht]
\centering
\caption{Runtime comparison across methods.}
\label{tab:pipeline_runtime}
\begin{tabular}{lccc}
\toprule
 Method & Runtime (min) \\
\midrule
LucidDreamer~\cite{chung2023luciddreamerdomainfreegeneration3d} & 8.04 \\
GenWarp~\cite{seo2024genwarpsingleimagenovel} & 0.62 \\
VistaDream~\cite{wang2024vistadreamsamplingmultiviewconsistent} & 24.70 \\
ViewCrafter~\cite{yu2024viewcraftertamingvideodiffusion} & 7.12 \\
Ours & 30.18 \\
\bottomrule
\end{tabular}
\end{table}

We provide a runtime comparison with baseline methods in Tab.~\ref{tab:pipeline_runtime}. Among the baselines, GenWarp~\cite{seo2024genwarpsingleimagenovel} and ViewCrafter~\cite{yu2024viewcraftertamingvideodiffusion} generate results significantly faster (0.62 and 7.12 minutes, respectively), primarily because they rotate the input view directly rather than synthesizing truly novel content under a fixed-camera 360° rotation trajectory. LucidDreamer~\cite{chung2023luciddreamerdomainfreegeneration3d} and VistaDream~\cite{wang2024vistadreamsamplingmultiviewconsistent} require longer runtimes (8.04 and 24.70 minutes, respectively) and are capable of full-scene generation; however, both methods exhibit notable shortcomings in terms of appearance consistency and geometric plausibility. Specifically, LucidDreamer~\cite{chung2023luciddreamerdomainfreegeneration3d} frequently produces abrupt style shifts and highly irregular 3D geometries across viewpoints, while VistaDream~\cite{wang2024vistadreamsamplingmultiviewconsistent} achieves slightly improved results but still fails to meet the standard of coherent scene generation. In contrast, our method AnchoredDream completes the full pipeline in approximately 30 minutes, generating a visually consistent and geometrically plausible 360° indoor scene. Although somewhat slower than certain baselines, AnchoredDream achieves a significantly better trade-off between efficiency and generation quality. Furthermore, our pipeline operates in a fully zero-shot manner, making it practical and scalable for real-world \noindent\mbox{deployment without substantial computational cost.}

\section{Additional Visual Results}

In this section, we present additional visual results to further illustrate the performance differences between our approach and the baselines.

Rather than exhaustively presenting every scene in the dataset, we curate a set of representative examples that span diverse appearance conditions and geometries. These examples reliably capture the typical performance differences between our method and the baselines. Including additional cases would lead to largely redundant visualizations, as the same qualitative patterns repeatedly emerge across the full evaluation set. The selected examples therefore provide a concise yet comprehensive summary, clearly illustrating the characteristic failure modes of baseline methods and the robustness and consistency of our approach.

\begin{figure}[ht]
  \centering
  \includegraphics[width=0.98\linewidth]{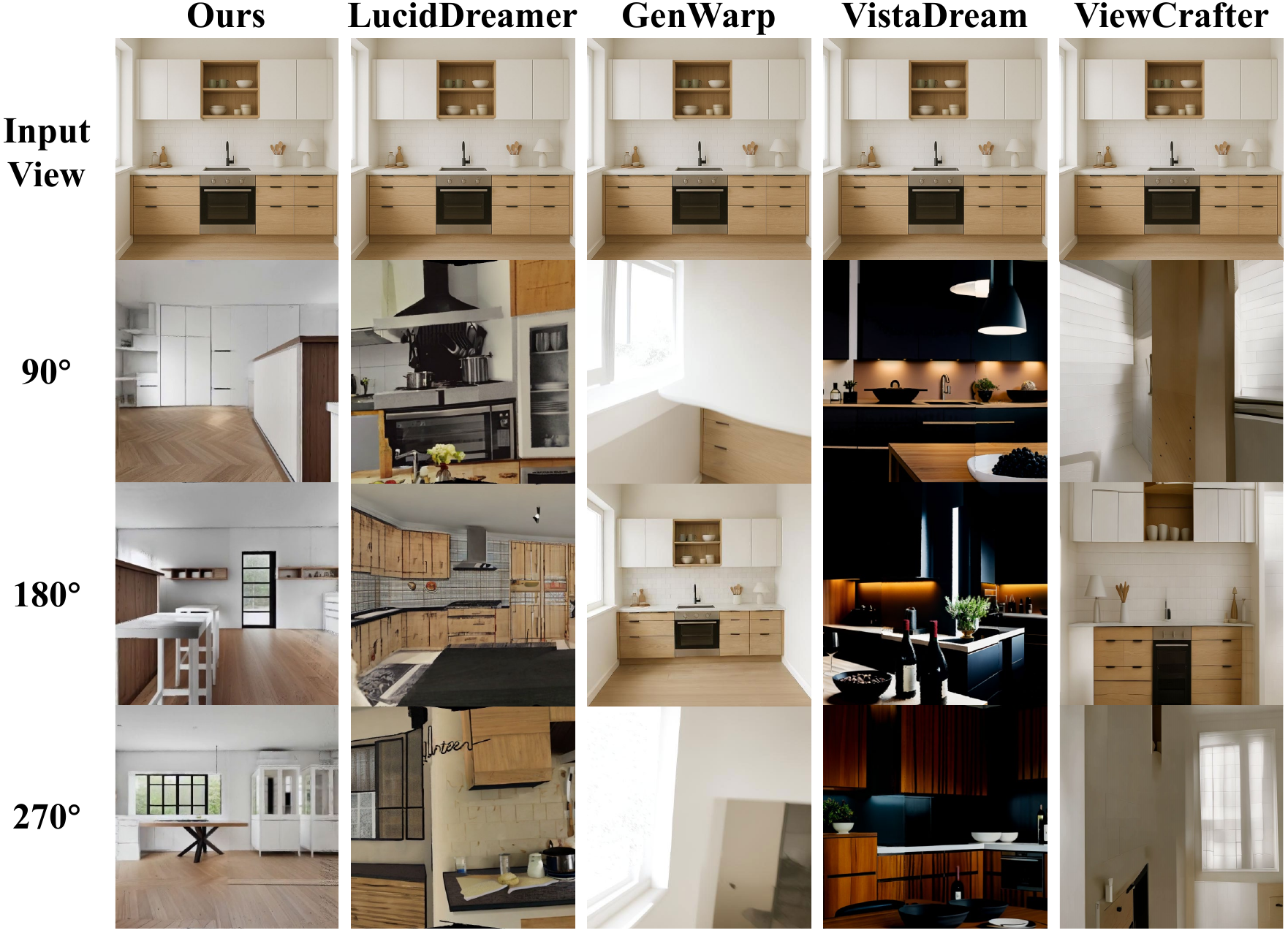}
\end{figure}

\begin{figure}[ht]
  \centering
  \includegraphics[width=0.98\linewidth]{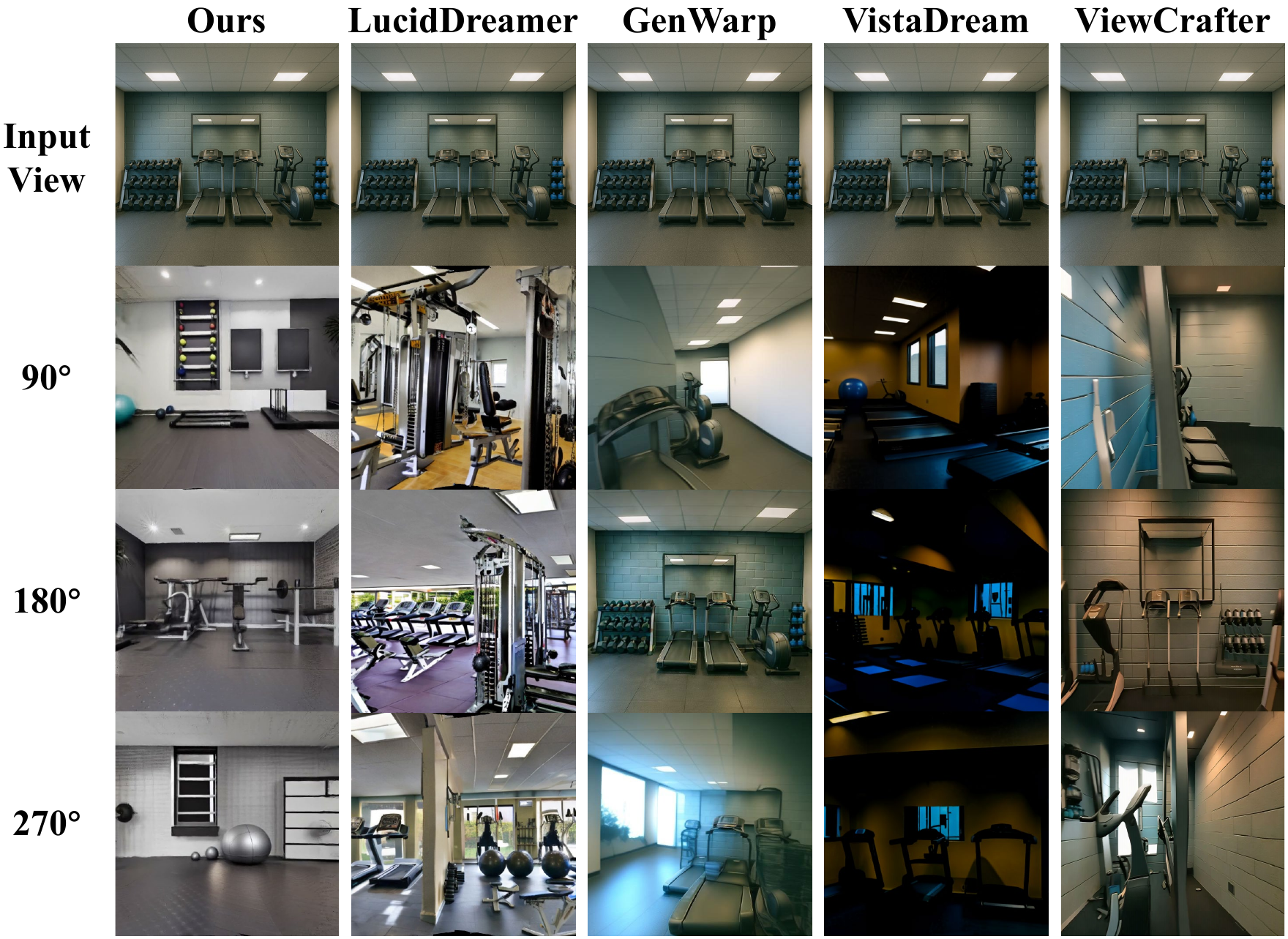}
\end{figure}

\begin{figure}[ht]
  \centering
  \includegraphics[width=0.98\linewidth]{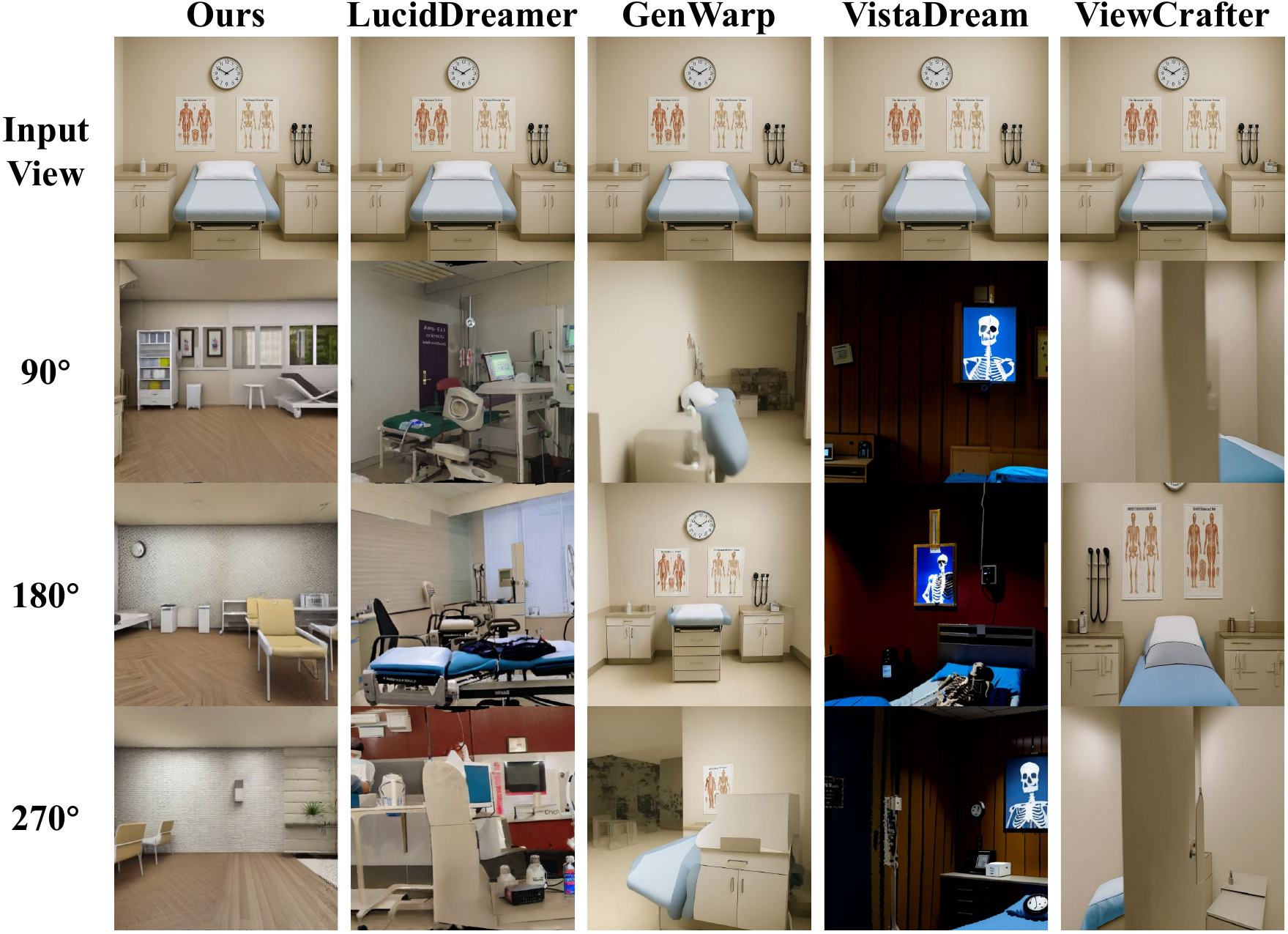}
\end{figure}

\begin{figure}[ht]
  \centering
  \includegraphics[width=\linewidth]{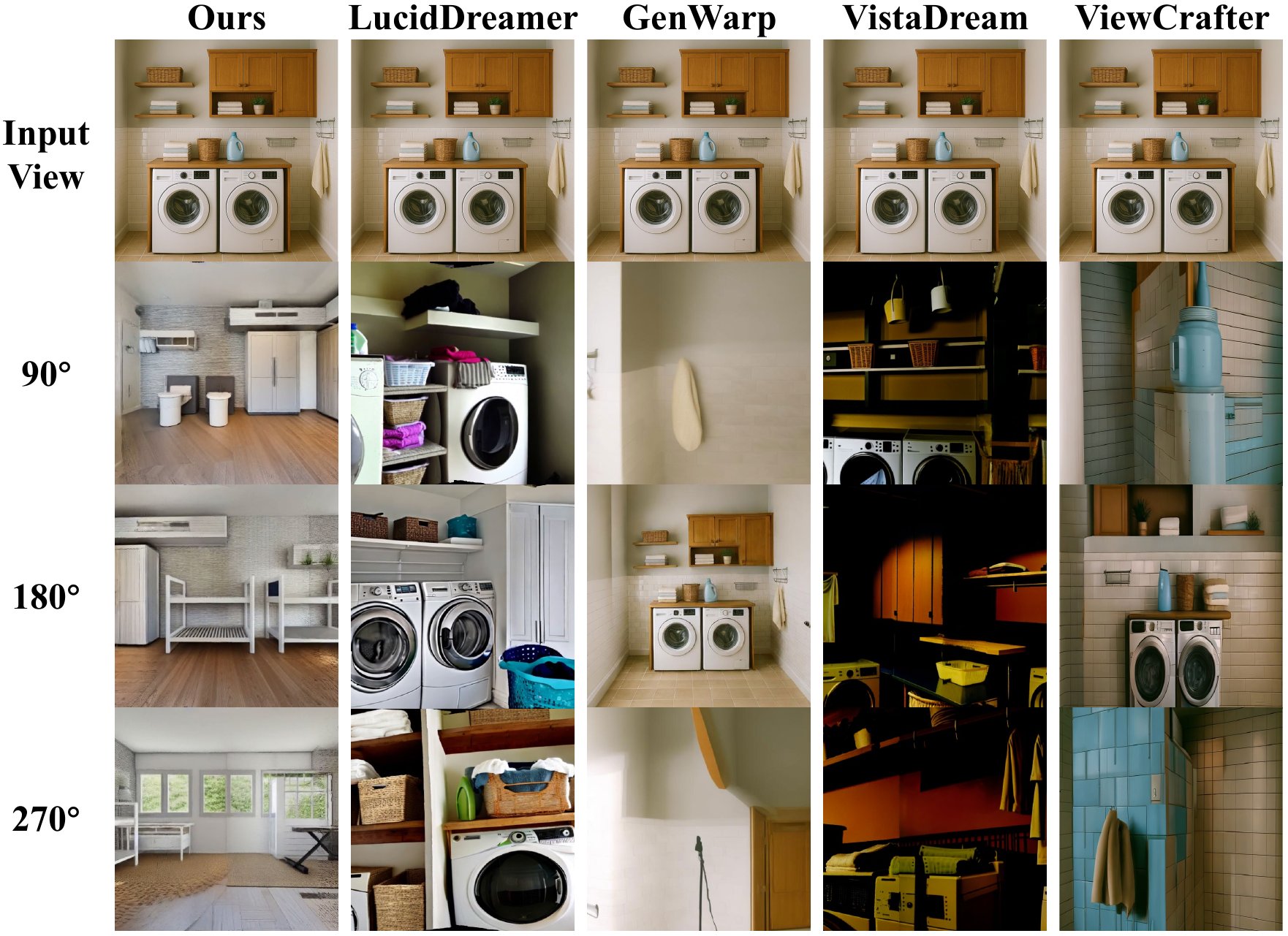}
\end{figure}

\begin{figure}[ht]
  \centering
  \includegraphics[width=\linewidth]{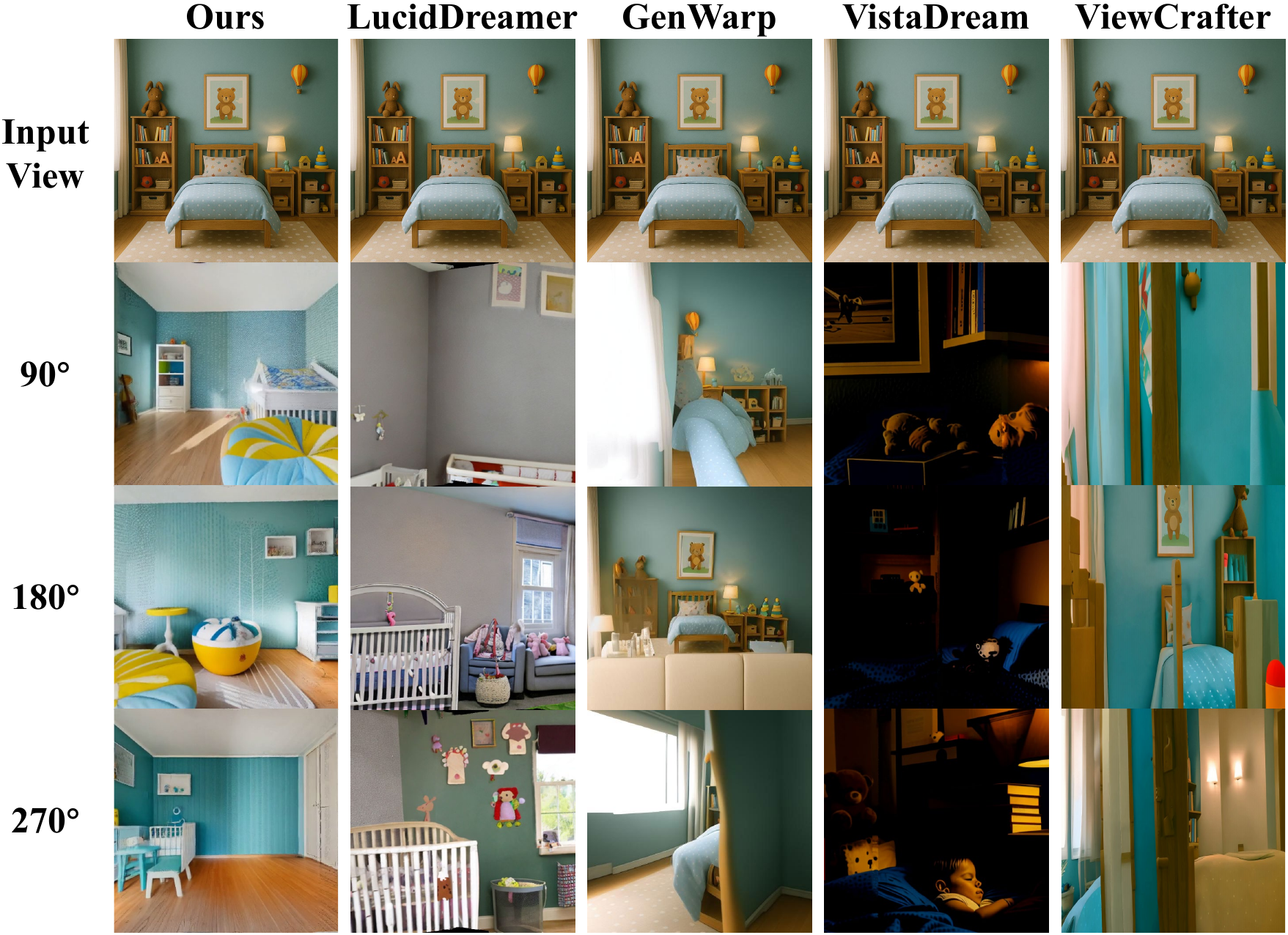}
\end{figure}

\begin{figure}[ht]
  \centering
  \includegraphics[width=\linewidth]{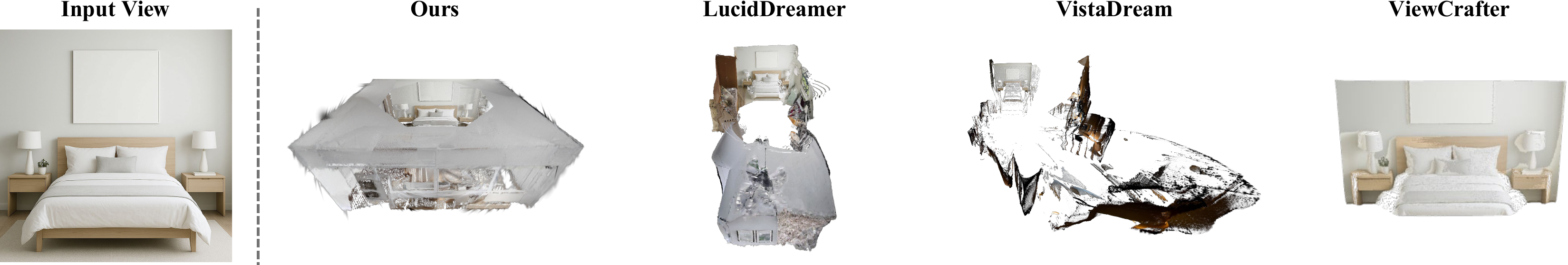}
\end{figure}

\begin{figure}[ht]
  \centering
  \includegraphics[width=\linewidth]{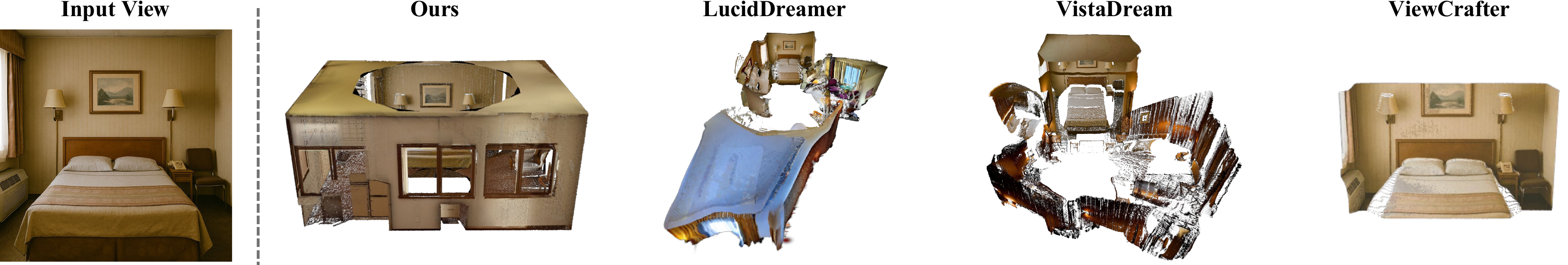}
\end{figure}

\begin{figure}[ht]
  \centering
  \includegraphics[width=\linewidth]{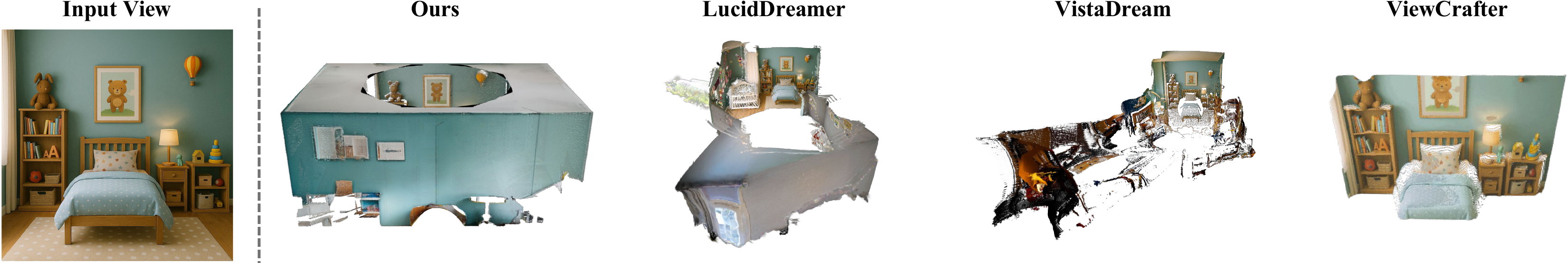}
\end{figure}

\end{document}